\documentclass[5p,times,twocolumn]{elsarticle}

\usepackage{amsmath,amssymb,amsfonts,color}
\usepackage{graphicx}
\usepackage{textcomp}
\usepackage{xcolor}
\usepackage{enumerate}
\usepackage{setspace}
\usepackage{bm}
\usepackage{comment}
\usepackage{caption}
\usepackage{tipa}
\usepackage{footnote}
\usepackage{subfigure}
\usepackage{algorithm}
\usepackage{algorithmic}
\usepackage{multirow}
\usepackage{multicol}
\usepackage{subfigure}

\def\x{{\mathbf x}}

\usepackage{hyperref}
\begin{document}

\begin{frontmatter}

\title{Importance-Aware Adaptive Dataset Distillation}

\author{Guang Li${}^\text{a}$}
\ead{guang@lmd.ist.hokudai.ac.jp}
\author{Ren Togo${}^\text{b}$}
\ead{togo@lmd.ist.hokudai.ac.jp}
\author{Takahiro Ogawa${}^\text{b}$}
\ead{ogawa@lmd.ist.hokudai.ac.jp}
\author{Miki Haseyama${}^\text{b}$}
\ead{mhaseyama@lmd.ist.hokudai.ac.jp}
\address{${}^\text{a}$Education and Research Center for Mathematical and Data Science, Hokkaido University, \\
           N-12, W-7, Kita-Ku, Sapporo, 060-0812, Japan}
\address{${}^\text{b}$Faculty of Information Science and Technology, Hokkaido University, \\
           N-14, W-9, Kita-Ku, Sapporo, 060-0814, Japan}

\begin{abstract}
Herein, we propose a novel dataset distillation method for constructing small informative datasets that preserve the information of the large original datasets.
The development of deep learning models is enabled by the availability of large-scale datasets.
Despite unprecedented success, large-scale datasets considerably increase the storage and transmission costs, resulting in a cumbersome model training process.
Moreover, using raw data for training raises privacy and copyright concerns.
To address these issues, a new task named dataset distillation has been introduced, aiming to synthesize a compact dataset that retains the essential information from the large original dataset.
State-of-the-art (SOTA) dataset distillation methods have been proposed by matching gradients or network parameters obtained during training on real and synthetic datasets.
The contribution of different network parameters to the distillation process varies, and uniformly treating them leads to degraded distillation performance.
Based on this observation, we propose an importance-aware adaptive dataset distillation (IADD) method that can improve distillation performance by automatically assigning importance weights to different network parameters during distillation, thereby synthesizing more robust distilled datasets.
IADD demonstrates superior performance over other SOTA dataset distillation methods based on parameter matching on multiple benchmark datasets and outperforms them in terms of cross-architecture generalization.
In addition, the analysis of self-adaptive weights demonstrates the effectiveness of IADD.
Furthermore, the effectiveness of IADD is validated in a real-world medical application such as COVID-19 detection.
\end{abstract}


\end{frontmatter}

\section{Introduction}
In the past few years, deep learning~\cite{schmidhuber2015deep, matsuo2022deep} has achieved tremendous success in diverse fields, including computer vision~\cite{krizhevsky2017imagenet}, natural language processing~\cite{young2018recent}, and speech recognition~\cite{amodei2016deep}.
The famous deep learning models such as AlexNet~\cite{krizhevsky2012imagenet}, ResNet~\cite{he2016deep}, BERT~\cite{devlin2019bert}, ViT~\cite{dosovitskiy2021image}, CLIP~\cite{radford2021learning}, Stable Diffusion~\cite{rombach2022high}, and ChatGPT~\cite{openai2023chatgpt} depend on large-scale datasets for training.
Nevertheless, the management of large-scale datasets presents a significant challenge, encompassing the intricate tasks of storage, transmission, and preprocessing~\cite{dai2019big}. 
For example, the challenge is underscored by the immense volume of data generated by high-resolution scans, such as MRI and CT scans, wherein a single 3D CT scan of the chest comprises thousands of individual slices, each containing millions of pixels~\cite{ernst2023sinogram}.
Moreover, to achieve suitable performance, training on large-scale datasets requires extensive computation, almost of the order of thousands of GPU hours~\cite{chen2021exploring, grill2020bootstrap, li2022tri}.
This problem can be alleviated by selecting representative training samples from large datasets, which is known as core-set or instance selection~\cite{goodfellow2013empirical, sener2017active}.
Some advanced instance selection methods have been proposed in recent years~\cite{mirzasoleiman2020coresets, killamsetty2021glister, killamsetty2021grad}.
Nevertheless, an upper bound exists on the compression rate achievable using data selection methods because certain original data cannot be discarded.
Furthermore, privacy and copyright issues are associated with the use of raw data for training~\cite{panch2018artificial, rieke2020future}.
\par
To address the aforementioned issues, a novel task named dataset distillation was proposed by Wang et al.~\cite{wang2018datasetdistillation} in 2018.
The objective of dataset distillation is to generate a small informative dataset ($\mathcal{D}_\textrm{distill}$) that allows models trained on it to perform comparably to those trained on the original dataset ($\mathcal{D}_\textrm{original}$), where the number of images in the synthetic dataset is considerably less than that in the original dataset.
Therefore, unlike dataset selection, dataset distillation method can compress a large dataset into a smaller dataset comprising only a few images, thereby substantially improving the dataset compression rate.
Because the generated distilled datasets do not contain raw data, the concerns of privacy and copyright can be addressed.
Downstream applications have substantially benefited from dataset distillation, such as continual learning~\cite{wiewel2021soft, sangermano2022sample}, privacy~\cite{dong2022privacy, chen2022privacy, wu2022towards, liu2023backdoor}, biomedical~\cite{li2020soft, li2022compressed, li2023sharing}, federated learning~\cite{song2023federated, xiong2022feddm, liu2023meta}, graph neural networks~\cite{jin2022graph, jin2022condensing}, neural architecture search~\cite{such2020generative, zhao2021datasetcondensation}, and black box optimization~\cite{chen2022bidirectional, chen2023bidirectional}.
\par
Dataset distillation has attracted increasing attention in recent years~\cite{sachdeva2023survey, lei2023survey, yu2023review}.
The original dataset distillation task is typically considered a meta-learning problem involving bi-level optimization~\cite{wang2018datasetdistillation}.
It optimizes the synthetic dataset by minimizing the classification loss on the original dataset in the outer loop and subsequently formulates the network parameters as functions of the learnable synthetic dataset in the inner loop.
Because gradient descent involves thousands to millions of steps, this recursive computation renders it unsuitable for real-world applications.
Zhao et al.~\cite{zhao2021datasetcondensation} proposed learning synthetic datasets by matching gradients or network parameters between real and synthetic data, while training teacher (using $\mathcal{D}_\textrm{original}$) and student (using $\mathcal{D}_\textrm{distill}$) networks to avoid unrolling the recursive computation graph.
Several methods have been subsequently proposed to improve dataset distillation performance via differentiable data augmentation~\cite{zhao2021differentiatble}, feature alignment~\cite{wang2022cafe, zhao2023distribution}, contrastive signaling~\cite{lee2022dataset}, and trajectory matching~\cite{cazenavette2022dataset, du2023minimizing}.
\begin{figure}[t]
        \centering
        \includegraphics[width=8.5cm]{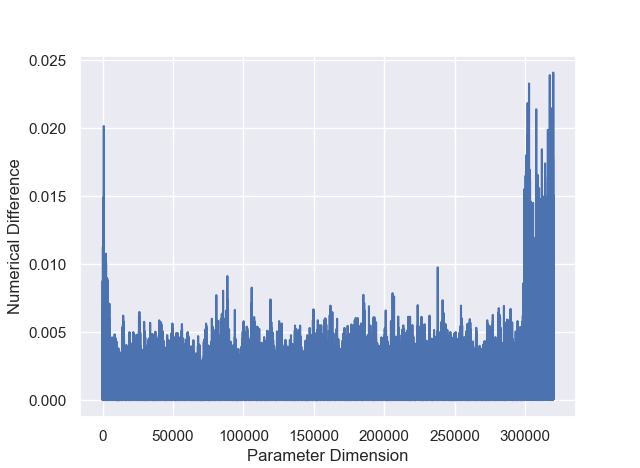}
        \caption{Visualization results of the numerical difference between parameters of teacher and student networks in the corresponding dimensions (A three-depth ConvNet on CIFAR-10).}
        \label{fig0}
\end{figure}
\par 
Although the concept of matching gradients or network parameters is intuitive and effective, a generally overlooked problem, i.e., whether each network parameter has the same importance in the dataset distillation process remains. 
Different network parameters exhibit different matching difficulties during the distillation process.
In particular, we distilled the CIFAR-10 dataset~\cite{krizhevsky2009learning} using a three-depth ConvNet containing 320,010 parameters.
For example, as depicted in Fig.~\ref{fig0}, the numerical difference between the teacher and student networks of a few network parameters is low, whereas others are high.
This implies that different network parameters differently contribute to the distillation process and equally treating these parameters affects the distillation performance.
Based on this observation, different network parameters can be attributed to different importance weights, and these weights can be subsequently optimized during the distillation process.  
Therefore, in this study, we explore this characteristic to further improve dataset distillation performance based on gradient (parameter) matching. 
\begin{figure}[t]
        \centering
        \includegraphics[width=8.5cm]{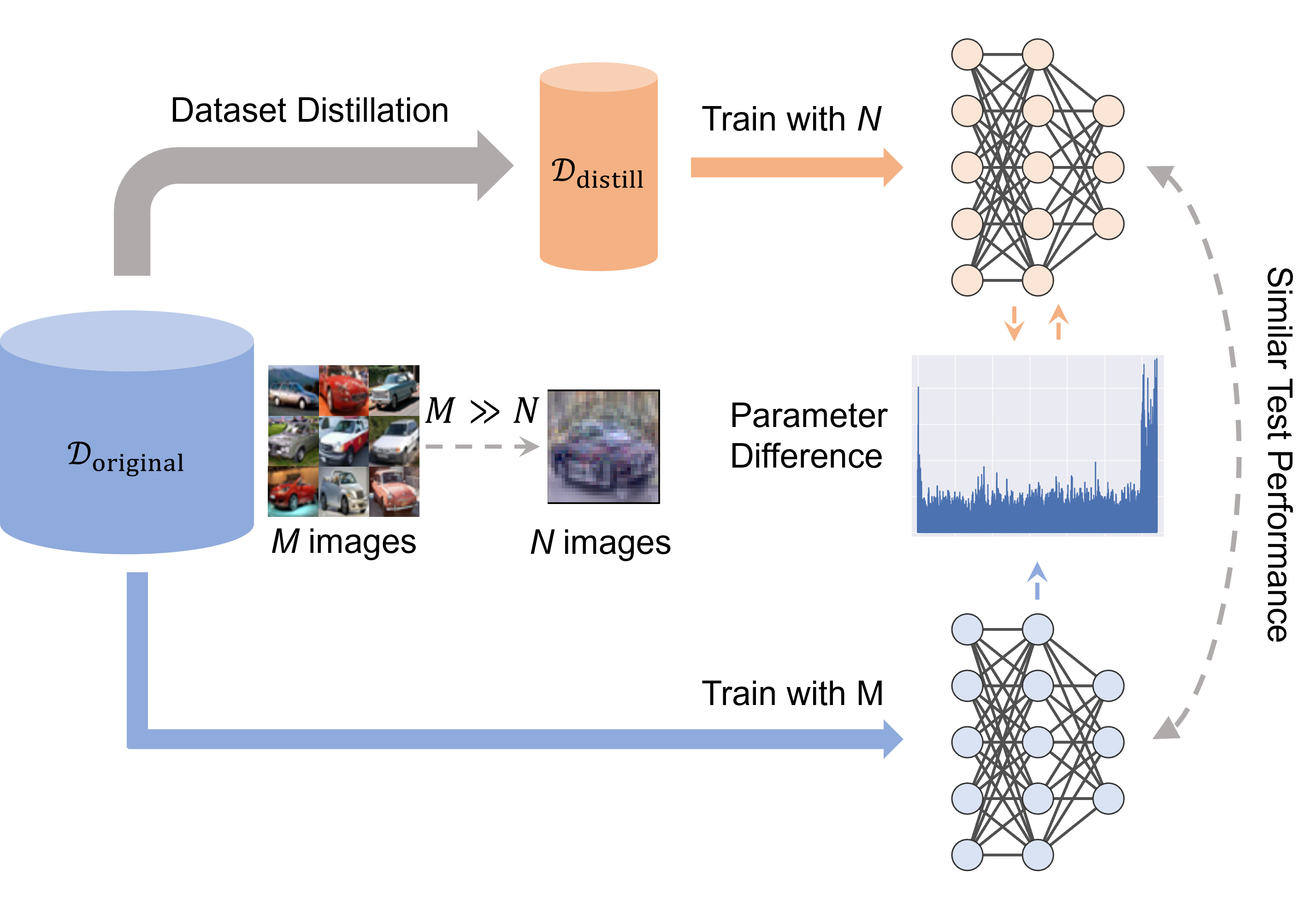}
        \caption{Concept of the proposed method. IADD aims to train the student network parameters by leveraging a distilled dataset that aligns with the teacher network parameters derived from the original large dataset. Because the parameter pairs in the teacher and student networks are different, we deal with these parameters using different importance weights.}
        \label{fig1}
\end{figure}
\par
Herein, we propose a novel dataset distillation method for solving the aforementioned above.
The concept of the proposed method is depicted in Fig.~\ref{fig1}.
Importance-aware adaptive dataset distillation (IADD) can improve distillation performance by automatically assigning importance weights to different network parameters and synthesizing more robust distilled datasets.
This ensures that crucial network parameters receive a higher weight, resulting in an improved contribution to the overall performance, whereas the effect of unimportant parameters is minimized. 
Furthermore, the optimization process is performed iteratively to refine the importance weights of network parameters until a desired level of performance is achieved.
Instead of equally treating all network parameters similar to conventional distillation methods, IADD considers the relative importance of each parameter and accordingly optimizes the importance weights.
This leads to more precise distillation and improved performance.
We performed extensive experiments in different settings to demonstrate the superiority of IADD.
\par
In summary, our study makes the following significant contributions:
\begin{itemize}
    \item We propose a novel dataset distillation method, IADD, for synthesizing small informative datasets that preserve the information of the large original datasets. 
    \item We design a framework that can automatically assign importance weights to different network parameters during distillation and synthesize more robust distilled datasets.
    \item We demonstrate that IADD achieves superior performance over other state-of-the-art (SOTA) dataset distillation methods based on parameter matching on multiple benchmark datasets and is more effective in cross-architecture generalization.
    \item We apply IADD to a real-world medical application such as COVID-19 detection.
\end{itemize}
\par
This study extends our prior research, dataset distillation using parameter pruning (DDPP), which primarily focused on parameter pruning~\cite{li2023ddpp}.
In this study, we augment our earlier work in the following aspects:
First, we introduced a novel dataset distillation approach capable of dynamically assigning adaptive weights to various parameters instead of directly eliminating them.
Second, we performed extensive experiments across four distinct datasets and varied settings to demonstrate the superior performance of IADD.
Finally, we visualized the parameter difference and self-adaptive weights and performed a detailed analysis to gain a more comprehensive understanding of the reasons underlying the improved performance of IADD.
\par
This paper is structured as follows.
Section 2 provides an overview of related works.
Section 3 details the proposed method.
Section 4 presents the experimental results and analysis.
Section 5 presents a discussion of our findings.
Finally, Section 6 provides the concluding remarks.
\section{Related Works}
\subsection{Dataset Distillation Based on Performance Matching}
In this section, we introduce dataset distillation methods based on performance matching.
Performance matching-based methods are designed to refine a distilled dataset to enable neural networks trained on it to achieve minimal loss when applied to the original dataset.
Consequently, this ensures matched performance between models trained using both the distilled and original datasets.
The first dataset distillation method was proposed by Wang et al.~\cite{wang2018datasetdistillation}.
In their method, model weights are defined as functions of distilled images and optimized using gradient-based hyperparameter optimization, which is widely used in meta-learning~\cite{finn2017model}.
Subsequently, certain methods extend the original method using flexible labels~\cite{bohdal2020flexible} or soft-label~\cite{sucholutsky2021soft} and add momentum~\cite{deng2022remember} to improve the distillation performance.
\par
\begin{figure*}[t]
        \centering
        \includegraphics[width=14cm]{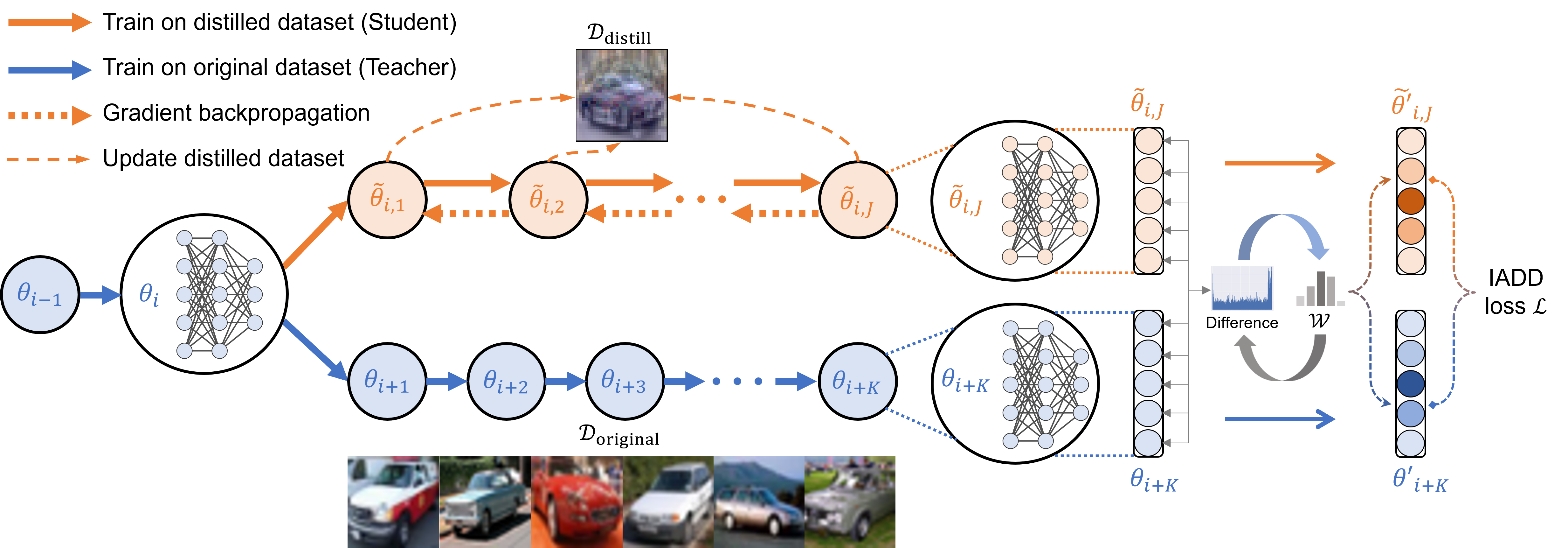}
        \caption{Overview of the proposed method. The main objective is to match the student network parameters $\tilde{\theta}'_{i,J}$ with the teacher network parameters $\theta'_{i+K}$ using the IADD loss $\mathcal{L}$. $\tilde{\theta}_{i,J}$ and $\theta_{i+K}$ denote intermediate model parameters without the processing of self-adaptive weights $\mathcal{W}$. $i$ represents the random start timestamp of the teacher and student parameters. $J$ and $K$ represent gradient descent updates of teacher and student parameters, respectively.}
        \label{fig2}
\end{figure*}
Backpropagation is used in meta-learning methods to calculate the gradient of validation loss on synthetic datasets, which necessitates bi-level optimization~\cite{maclaurin2015gradient, lorraine2020optimizing}.
Consequently, the optimization of outer loops necessitates extensive computing costs, and GPU memory requirements increase as the number of inner loops increases~\cite{vicol2022implicit}.
A limited number of inner loops results in insufficient inner optimization and performance bottlenecks, and upscaling this learning schema to large models is also inconvenient.
However, kernel inducing point (KIP), can solve this problem by performing convex optimization, resulting in a closed-form solution, avoiding the requirement for extensive inner loop training~\cite{nguyen2021kip}.
KIP involves initializing a labeled support set and iterating over a process where a random kernel and batches from the support and target dataset are sampled. It computes the kernel ridge-regression loss and updates the support set based on the loss.
KIP includes variations like randomly augmenting sampled target batches and introducing a corruption fraction to the data, resulting in highly corrupted yet effective datasets.
Several methods extended from KIP, which considerably improve distillation performance and efficiency, have been proposed such as infinitely wide convolutional networks~\cite{nguyen2021kipimprovedresults}, neural feature regression~\cite{zhou2022dataset}, and random feature approximation~\cite{loo2022efficient}.
\subsection{Dataset Distillation Based on Parameter Matching}
In this section, we illustrate dataset distillation methods based on parameter matching.
The first parameter matching method was proposed by Zhao et al.~\cite{zhao2021datasetcondensation}, which is also known as dataset condensation.
While performance matching focuses on optimizing the performance of networks trained on synthetic datasets, parameter matching optimizes the performance of networks trained on the original and synthetic datasets by ensuring the consistency of the trained network parameters.
Several studies have extended parameter matching using differentiable data augmentation~\cite{zhao2021differentiatble}, contrastive signaling~\cite{lee2022dataset}, long-range trajectory matching~\cite{cazenavette2022dataset}, and parameter pruning~\cite{li2023ddpp}.
\par
As noted in our earlier study DDPP~\cite{li2023ddpp}, some parameters are challenging to match in dataset distillation, resulting in a deteriorated distillation performance.
To address this issue, we used parameter pruning to eliminate such challenging parameters during the distillation process.
Nevertheless, removing these parameters is imprudent and can negatively influence the number of network model parameters.
Therefore, herein, we propose a novel method that can assign adaptive weights to different parameters instead of directly removing parameters to solve this problem.
\subsection{Dataset Distillation Based on Distribution Matching}
We present several dataset distillation methods based on distribution matching here.
Distribution matching produces synthetic data with a distribution close to those of the original data in a family of embedding spaces using the maximum mean discrepancy (MMD~\cite{gretton2012kernel}).
The first distribution matching method proposed by Zhao et al.~\cite{zhao2023distribution} used the output embeddings of neural networks without the last linear layer.
For each class of synthetic and original datasets, the mean vector (center) is preferred to be close.
Subsequently, another method has been proposed to force the statistics of features extracted by each network layer to be consistent~\cite{wang2022cafe}.
Although these methods reduce the synthesis cost and are generally applicable to larger datasets, the distillation performance is not improved.
For more details regarding the literature on dataset distillation, please refer to the recent survey papers~\cite{sachdeva2023survey, lei2023survey, yu2023review} or Awesome-Dataset-Distillation~\cite{li2022awesome}.
\subsection{Dataset Distillation for Medical Task}
Dataset distillation, which can reduce the complexity and volume of medical data improves efficiency in medical tasks.
It has been applied to several medical scenarios, such as gastritis detection~\cite{li2020soft}, COVID-19 detection~\cite{li2023sharing}, skin lesion classification~\cite{tian2023gdd}, and medical data sharing~\cite{li2022compressed}.
However, in the dynamic field of medical applications, diverse innovative approaches have emerged.
For example, hybrid solutions combine conventional machine learning, deep learning, and domain-specific knowledge, creating a synergy of methodologies~\cite{wang2020deep, jared2022inter}. 
These methods improve interpretability, while utilizing the predictive potential of deep learning, which is particularly valuable for medical diagnosis and risk assessment~\cite{gavrishchaka2019synergy}.
Inspired by neuroscience principles, neuroheuristics have made notable strides in medical applications~\cite{ke2019neuro, rajagopal2023deep}. 
These methods emulate neural processes, thereby improving the comprehension and analysis of intricate medical data. 
Neural networks designed to mirror neural functionality have improved diagnostic accuracy and decision support systems~\cite{si2019enhancing}. 
Multimodal fusion methods play a crucial role in integrating diverse data sources, including medical imaging, electronic health records, and genomic data~\cite{wang2019ai, yang2022unbox}.
This integration provides valuable insights crucial for personalized medicine and disease prediction in an era marked by the prevalence of diverse medical data types~\cite{subramanian2020precision, tran2021deep}.
Therefore, recognizing the importance of dataset distillation methods in the medical domain and contemplating their potential synergy and collaboration with other methodologies is imperative.
\section{Methodology}
Figure~\ref{fig2} provides an overview of the proposed method.
IADD aims to train the student network parameters by leveraging a distilled dataset $\mathcal{D}_\textrm{distill}$ that aligns with the teacher network parameters derived from the original large dataset $\mathcal{D}_\textrm{original}$.
The proposed method involves three main stages: (1) training the teacher and student networks, (2) matching teacher and student parameters, and (3) generating an optimized distilled dataset.
\begin{algorithm}[t]
    \caption{Importance-Aware Adaptive Dataset Distillation}
    \label{alg1}
    \begin{algorithmic}[1]
    \REQUIRE 
    $\{\theta_{i}\}^{I}_{0}$: teacher parameters trained on $\mathcal{D}_\textrm{original}$;
    $\mathcal{W}_{0}$: initial value for $\mathcal{W}$;
    $\alpha_{0}$: initial value for $\alpha$;
    $\mathcal{A}$: differentiable augmentation function;
    $\epsilon$: threshold for pruning;
    $T$: number of distillation iterations;
    $J$: number of updates for the student network;
    $K$: number of updates for the teacher network;
    $I^{+}$: upper bound of the sample range of $i$.
    \ENSURE
    optimized trainable learning rate $\alpha^{\ast}$,
    optimized self-adaptive weights $\mathcal{W}^{\ast}$, and
    optimized distilled dataset $\mathcal{D}^{\ast}_\textrm{distill}$.
    \\
    \STATE
    Initialize distilled dataset:
    $\mathcal{D}_\textrm{distill} \thicksim \mathcal{D}_\textrm{original}$
    \STATE
    Initialize trainable learning rate:
    $\alpha = \alpha_{0}$
    \FOR{each distillation iteration $t = 0$ to $T - 1$}
    \STATE
    Select random start timestamp $i < I^{+}$
    \STATE
    Initialize student network using teacher parameter: 
    $\tilde{\theta}_{i}=\theta_{i}$
    \FOR{each student update $j = 0$ to $J - 1$}
    \STATE
    Sample a minibatch of the distilled dataset:
    $b_{i,j} \thicksim \mathcal{D}_\textrm{distill}$ 
    \STATE
    Update student network using the cross-entropy loss $\ell$:
    \STATE
    $\tilde{\theta}_{i,j+1} = \tilde{\theta}_{i,j} - \alpha\nabla_{\tilde{\theta}_{i,j}}\ell(\mathcal{A}(b_{i,j}))$
    \ENDFOR
    \STATE
    Obtain importance-aware parameters $\tilde{\theta}'_{i,J}$ and $\theta_{i+K}'$ with Eqs. (3)--(8)
    \STATE
    Calculate the IADD loss:
    \STATE
    $\mathcal{L} = || \tilde{\theta}'_{i,J}-\theta'_{i+K} ||^{2}_{2} \,\,\,/\,\,\, || \theta_{i}-\theta_{i+K} ||^{2}_{2}$
    \STATE
    Update $\alpha$, $\mathcal{W}$, and $\mathcal{D}_\textrm{distill}$ with with Eqs. (10)--(13)
    \ENDFOR
    \end{algorithmic}
\end{algorithm}
\subsection{Teacher and Student Network Training}
In the proposed method, $N$ teacher networks are first pretrained on the original dataset $\mathcal{D}_\textrm{original}$ for the distillation process, and their snapshot parameters are saved at each epoch.
Note that pretraining of many teacher networks is only performed once for one dataset and can be reused.
To improve the robustness and diversity of knowledge transfer during the distillation process, we use multiple teacher networks. 
Each teacher network represents a distinct snapshot of the model during training, capturing unique aspects of the data and the learning process. 
These snapshots collectively provide a more comprehensive understanding of the intricacies of the dataset and model training process.
By utilizing multiple teacher networks, we can reduce the risk of overfitting to a single snapshot or encountering biases in a particular training stage.
This ensemble of teacher networks contributes to the stability and reliability of the dataset distillation process, ultimately enhancing the generalization ability and effectiveness of IADD.
Initialization of the distilled is performed using random samples from the original dataset as $\mathcal{D}_\textrm{distill} \thicksim \mathcal{D}_\textrm{original}$.
The number of images in the distilled dataset can be obtained by IPC, where IPC denotes the number of distilled images per class and is typically used in the dataset distillation task.
The teacher parameters are defined as a sequence of parameters $\{\theta_{i}\}^{I}_{0}$, where $I$ represents the number of training steps.
To initiate student parameters, a teacher parameter is selected at a random start timestamp $i$.
The student parameters are initialized as $\tilde{\theta}_{i}=\theta_{i}$.
To avoid using less informative parts of the teacher parameters, we introduce an upper bound $I^{+}$ on the random start timestamp $i$.
Updating the student and teacher parameters involves executing $J$ and $K$ gradient descent updates, respectively, where $J \ll K$.
For each student update $j$, a minibatch $b_{i,j}$ is sampled from the distilled dataset $\mathcal{D}_\textrm{distill}$ as follows:
\begin{equation}
b_{i,j} \thicksim \mathcal{D}_\textrm{distill}.
\end{equation}
Subsequently, the student parameters $\tilde{\theta}$ undergo $j$ updates using the cross-entropy loss $\ell$ to optimize performance as follows: 
\begin{equation}
\tilde{\theta}_{i,j+1} = \tilde{\theta}_{i,j} - \alpha\nabla_{\tilde{\theta}_{i,j}} \ell (\mathcal{A}(b_{i,j})),
\end{equation}
where $\alpha$ represents a trainable learning rate that can be updated during the training process.
Moreover, the differentiable data augmentation module $\mathcal{A}$, as proposed in~\cite{zhao2021differentiatble}, is used to improve the distillation performance.
In particular, the loss $\ell$ is computed as the average of the cross-entropy between the predictions of the student network and the corresponding one-hot labels. 
And $\nabla_{\tilde{\theta}_{i,j}} \ell$ denotes the gradients of $\ell$ based on $\tilde{\theta}_{i,j}$.
The updates of student parameters are performed via stochastic gradient descent (SGD).
\subsection{Teacher and Student Parameters Matching}
Once $J$ updates are performed, the student parameters $\tilde{\theta}_{i,J}$ can be obtained, which are then trained using the distilled dataset $\mathcal{D}_\textrm{distill}$ with the student network initialized.
Similarly, we obtain the teacher parameters $\theta_{i+K}$ trained on the original dataset $\mathcal{D}_\textrm{original}$ with $K$ updates, which are saved in the pretrained snapshots.
Subsequently, the initiated parameters $\theta_{i}$, student parameters $\tilde{\theta}_{i,J}$, and teacher parameters $\theta_{i+K}$ are converted into 1D vectors by concatenating the flattened parameters of each layer as follows:
\begin{equation}
\theta_{i} = [\theta_{i}^{1}, \theta_{i}^{2},\dotsb,\theta_{i}^{P}],
\end{equation}
\begin{equation}
\tilde{\theta}_{i,J} = [\tilde{\theta}_{i,J}^{1}, \tilde{\theta}_{i,J}^{2},\dotsb,\tilde{\theta}_{i,J}^{P}],
\end{equation}
\begin{equation}
\theta_{i+K} = [\theta_{i+K}^{1}, \theta_{i+K}^{2},\dotsb,\theta_{i+K}^{P}].
\end{equation}
Here, $P$ denotes the total number of parameters.
The transformation creates a unified parameter vector by aggregating parameters from different layers. 
The flattened parameters streamline the subsequent operations such as parameter matching and weight allocation, facilitating the model alignment and dataset distillation processes.
We define the self-adaptive weights and importance-aware parameters as follows:
\begin{equation}
\mathcal{W} = [w^{1}, w^{2},\dotsb, w^{P}],
\end{equation}
\begin{equation}
\tilde{\theta}'_{i,J} = \mathcal{W}\tilde{\theta}_{i,J} = [w^{1}\tilde{\theta}_{i,J}^{1}, w^{2}\tilde{\theta}_{i,J}^{2},\dotsb,w^{P}\tilde{\theta}_{i,J}^{P}],
\end{equation}
\begin{equation}
\theta_{i+K}' = \mathcal{W}\theta_{i+K} = [w^{1}\theta_{i+K}^{1}, w^{2}\theta_{i+K}^{2},\dotsb,w^{P}\theta_{i+K}^{P}],
\end{equation}
The self-adaptive weights $\mathcal{W}$ are initialized with $\mathcal{W}_{0}$, a unit vector with corresponding dimensions.
The IADD loss $\mathcal{L}$ calculates the squared $L_{2}$ error between the importance-aware student parameters $\tilde{\theta}'_{i,J}$ and teacher parameters $\theta'_{i+K}$ as follows:
\begin{equation}
\mathcal{L} = \frac{|| \tilde{\theta}'_{i,J}-\theta'_{i+K} ||^{2}_{2}} {|| \theta_{i}-\theta_{i+K} ||^{2}_{2}}.
\end{equation}
where the distance between the teacher parameters at starting and ending updates $\theta_{i}\,-\,\theta_{i+K}$, ensuring that the late training stage of teacher networks can still provide sufficient supervision, even if it has converged, thereby allowing for a more effective comparison between the student and teacher parameters.
Automatically assigning importance weights to different network parameters during distillation can improve the contribution of essential parameters and penalize unimportant parameters, improving the distillation performance and generating more robust distilled datasets.
\subsection{Optimized Distilled Dataset Generation}
Finally, we minimize the IADD loss $\mathcal{L}$ with updates on sampled distilled dataset $\mathcal{A}(b_{i,j})$ via SGD.
Subsequently, we optimize the trainable learning rate $\alpha$, self-adaptive weights $\mathcal{W}$, and distilled dataset $\mathcal{D}_\textrm{distill}$ through all $J$ updates of the student network as follows:
\begin{equation}
\alpha_{i,j+1} = \alpha_{i,j} - \mu \nabla_{\alpha_{i,j}} \mathcal{L}(\mathcal{A}(b_{i,j})),
\end{equation}
\begin{equation}
\mathcal{W}_{i,j+1} = \mathcal{W}_{i,j} - \eta \nabla_{\mathcal{W}_{i,j}} \mathcal{L}(\mathcal{A}(b_{i,j})),
\end{equation}
\begin{equation}
\mathcal{D}_{\textrm{distill}_{i,j+1}} = \mathcal{D}_{\textrm{distill}_{i,j}} - \zeta \nabla_{\mathcal{D}_{\textrm{distill}_{i,j}}} \mathcal{L}(\mathcal{A}(b_{i,j})),
\end{equation}
where $\nabla_{\alpha_{i,j}} \mathcal{L}$, $\nabla_{\mathcal{W}_{i,j}} \mathcal{L}$ and $\nabla_{\mathcal{D}_{\textrm{distill}_{i,j}}} \mathcal{L}$ represent the gradients of $\mathcal{L}$ based on $\alpha_{i,j}$, $\mathcal{W}_{i,j}$ and $\mathcal{D}_{\textrm{distill}_{i,j}}$, respectively.
$\mu$, $\eta$, and $\zeta$ represent the corresponding learning rates.
The self-adaptive weights $\mathcal{W}$ are updated to optimize the IADD process, which can improve the performance of the distilled network.
The generated distilled dataset is then used to train the student network in the next iteration, and this process continues until convergence.
Finally, we can obtain the optimized learning rate $\alpha^{\ast}$, self-adaptive weights $\mathcal{W}^{\ast}$, and distilled dataset $\mathcal{D}^{\ast}_\textrm{distill}$ as follows:
\begin{equation}
\alpha^{\ast}, \mathcal{W}^{\ast}, \mathcal{D}^{\ast}_\textrm{distill} = \mathrm{arg \, min} \,\mathcal{L} (\alpha, \mathcal{W}, \mathcal{D}_\textrm{distill}; \tilde{\theta}).
\end{equation}
The distillation process is summarized in Algorithm~\ref{alg1}.
\par
The optimized learning rate $\alpha^{\ast}$ can serve as an adaptive regulator for the number of updates performed on the student and teacher networks ($J$ and $K$, respectively), simplifying the distillation process optimization.
The optimized self-adaptive weights $\mathcal{W}^{\ast}$ maximize the contribution of parameter matching of different importance to dataset distillation, resulting in higher precise distillation and improved performance.
\par
Once we obtain the optimized distilled dataset $\mathcal{D}^{\ast}_\textrm{distill}$, we can use it to train various networks.
Because of the significantly reduced volume of distilled data compared with the original data, the distilled dataset offers notable benefits. 
For example, it leads to substantial savings in data storage space, which is particularly advantageous when dealing with large-scale datasets.
Furthermore, the streamlined dataset size increases the model training speed, thereby reducing computational resource requirements and allowing for more rapid model iteration.
In addition, the distilled dataset shows potential for diverse downstream applications. 
In particular, in contexts where data privacy is of utmost importance, such as in the medical imaging field and other sensitive domains, the distilled dataset can play a pivotal role.
\section{Experiments}
In this section, we conduct extensive experiments in different settings to demonstrate the superiority of IADD.
We have provided the details of our experimental settings in Section 4.1.
Sections 4.2, 4.3, 4.4, and 4.5 report the benchmark comparison, cross-architecture generalization, analysis of self-adaptive weights, and real-world medical application, respectively.
\subsection{Experimental Settings}
\begin{table*}[t]
    \centering
    \caption{Test accuracy compared with SOTA dataset distillation methods on several benchmark datasets. IPC represents the number of distilled images per class. The first three comparison methods are instance selection methods, and the others are SOTA dataset distillation methods. All the presented results are the average value obtained over five trials.}
    \label{tab1}
    \begin{tabular}{cc|ccc|ccc|cc}
    \hline
    \multicolumn{2}{c|}{Dataset} &\multicolumn{3}{c|}{CIFAR-10} &\multicolumn{3}{c|}{CIFAR-100} &\multicolumn{2}{c}{Tiny ImageNet} \\
    \multicolumn{2}{c|}{IPC} & 1 & 10 & 50 & 1 & 10 & 50 & 1 & 10 \\\hline\hline
    \multicolumn{2}{c|}{Random} 
    & 14.4$\pm$2.0 & 26.0$\pm$1.2 & 43.4$\pm$1.0 & 4.2$\pm$0.3 & 14.6$\pm$0.5 & 30.0$\pm$0.4 & 1.4$\pm$0.1 & 5.0$\pm$0.2 \\
    \multicolumn{2}{c|}{Forgetting} 
    & 13.5$\pm$1.2 & 23.3$\pm$1.0 & 23.3$\pm$1.1 & 4.5$\pm$0.2 & 15.1$\pm$0.3 & 30.5$\pm$0.3 & 1.6$\pm$0.1 & 5.1$\pm$0.2 \\
    \multicolumn{2}{c|}{Herding} 
    & 21.5$\pm$1.2 & 31.6$\pm$0.7 & 40.4$\pm$0.6 & 8.4$\pm$0.3 & 17.3$\pm$0.3 & 33.7$\pm$0.5 & 2.8$\pm$0.2 & 6.3$\pm$0.2 \\\hline\hline
    \multicolumn{2}{c|}{DC~\cite{zhao2021datasetcondensation}} 
    & 28.3$\pm$0.5 & 44.9$\pm$0.5 & 53.9$\pm$0.5 & 12.8$\pm$0.3 & 25.2$\pm$0.3 & 29.7$\pm$0.3 & 5.3$\pm$0.2 & 11.1$\pm$0.3 \\
    \multicolumn{2}{c|}{DSA~\cite{zhao2021differentiatble}}
    & 28.8$\pm$0.7 & 52.1$\pm$0.5 & 60.6$\pm$0.5 & 13.9$\pm$0.3 & 32.3$\pm$0.3 & 42.8$\pm$0.4 & 6.6$\pm$0.2 & 16.3$\pm$0.2 \\
    \multicolumn{2}{c|}{DM~\cite{zhao2023distribution}}
    & 26.0$\pm$0.8 & 48.9$\pm$0.6 & 63.0$\pm$0.4 & 11.4$\pm$0.3 & 29.7$\pm$0.3 & 43.6$\pm$0.4 & 3.9$\pm$0.2 & 13.5$\pm$0.3 \\
    \multicolumn{2}{c|}{CAFE~\cite{wang2022cafe}}
    & 30.3$\pm$1.1 & 46.3$\pm$0.6 & 55.5$\pm$0.6 & 12.9$\pm$0.3 & 27.8$\pm$0.3 & 37.9$\pm$0.3 & - & - \\
    \multicolumn{2}{c|}{CAFE + DSA~\cite{wang2022cafe}}
    & 31.6$\pm$0.8 & 50.9$\pm$0.5 & 62.3$\pm$0.4 & 14.0$\pm$0.3 & 31.5$\pm$0.2 & 42.9$\pm$0.2 & - & - \\
    \multicolumn{2}{c|}{EGM~\cite{jiang2022delving}}
    & 30.0$\pm$0.6 & 50.2$\pm$0.6 & 60.0$\pm$0.4 & 12.7$\pm$0.4 & 31.1$\pm$0.3 & - & - & - \\
    \multicolumn{2}{c|}{DCC~\cite{lee2022dataset}}
    & 34.0$\pm$0.7 & 54.5$\pm$0.5 & 64.2$\pm$0.4 & 14.6$\pm$0.3 & 33.5$\pm$0.3 & 40.0$\pm$0.3 & - & - \\
    \multicolumn{2}{c|}{MTT~\cite{cazenavette2022dataset}}
    & 46.3$\pm$0.8 & 65.3$\pm$0.7 & 71.6$\pm$0.2 & 24.3$\pm$0.3 & 40.1$\pm$0.4 & 47.7$\pm$0.2 & 8.8$\pm$0.3 & 23.2$\pm$0.2 \\
    \multicolumn{2}{c|}{DDPP~\cite{li2023ddpp}}
    & 46.4$\pm$0.6 & 65.5$\pm$0.3 & 71.9$\pm$0.2 & 24.6$\pm$0.1 & \bfseries{43.1$\pm$0.3} & 48.4$\pm$0.3 & - & - \\
    \multicolumn{2}{c|}{IADD}
    & \bfseries{46.5$\pm$1.1} & \bfseries{66.7$\pm$0.8} & \bfseries{72.6$\pm$0.3} & \bfseries{25.2$\pm$0.1} & 42.7$\pm$0.5 & \bfseries{49.0$\pm$0.3} & \bfseries{9.6$\pm$0.4} & \bfseries{24.1$\pm$0.3} \\\hline\hline
    \multicolumn{2}{c|}{Original Dataset}
    & & 84.8$\pm$0.1 & & & 56.2$\pm$0.3 & & \multicolumn{2}{c}{37.6$\pm$0.4} \\\hline
    \end{tabular}
\end{table*}
\begin{table}[t]
    \centering
    \caption{Test accuracy compared with KIP~\cite{nguyen2021kipimprovedresults}.}
    \label{tab2}
    \begin{tabular}{l|c|c|cc}
    \hline
    & IPC & KIP-1024 & KIP-128 & IADD-128 \\\hline\hline
    \multirow{3}*{CIFAR-10} & 1 & 49.9 & 38.3 & \bfseries{46.5} \\
    & 10 & 62.7 & 57.6 & \bfseries{66.7} \\
    & 50 & 68.6 & 65.8 & \bfseries{72.6} \\\hline
    \multirow{3}*{CIFAR-100} & 1 & 15.7 & 18.2 & \bfseries{25.2} \\
    & 10 & 28.3 & 32.8 & \bfseries{42.7} \\
    & 50 & - & - & \bfseries{49.0} \\\hline
    \end{tabular}
\end{table}
Herein, three benchmark datasets, namely, CIFAR-10~\cite{krizhevsky2009learning}, CIFAR-100~\cite{krizhevsky2009learning}, and Tiny ImageNet~\cite{le2015tiny}, were used for the dataset distillation task.
The CIFAR-10 and CIFAR-100 datasets comprise images of size 32 $\times$ 32, whereas the Tiny ImageNet dataset comprises images of size 64 $\times$ 64.
The CIFAR-10 dataset comprises 60,000 color images with 10 classes. 
Within each class, there are 6,000 images, which are further divided into 50,000 training and 10,000 test images. 
The CIFAR-100 dataset mirrors the CIFAR-10 dataset in structure, but offers a more extensive variety with 100 classes.
It contains 60,000 color images with each class comprising 600 images.
The training set includes 500 images per class, and the test set includes 100 images per class. 
The Tiny ImageNet dataset comprises 200 training classes with each class comprising 500 color images. 
In addition to the training set, a test set is present, containing 10,000 images with 200 classes.
\par
In our experiments, we evaluated the proposed method on a large COVID-19 CXR dataset~\cite{rahman2021exploring} to demonstrate its effectiveness in real-world medical applications, namely, COVID-19 detection\footnote{https://www.kaggle.com/datasets/tawsifurrahman/covid19-radiography-database}.
The dataset comprises four classes: COVID-19 with 3,616 images, Lung Opacity with 6,012 images, Normal with 10,192 images, and Viral Pneumonia with 1,345 images.
The split ratio of the training and test sets is 8:2 for each class, resulting in 16,933 images for training and 4,232 images for testing.
We resized the CXR images from the original 224 $\times$ 224 to 112 $\times$ 112 for the dataset distillation process.
The resolution of images in the datasets was selected to ensure that the experiments were performed on datasets with varying complexities.
The training set of each dataset is used in the distillation process to generate the distilled data, and the test phase is performed on the test set of the original datasets by the student network.
\par
Extensive experiments were performed to evaluate the effectiveness of the proposed method across various settings, including benchmark comparison, cross-architecture generalization, self-adaptive weight analysis, and real-world medical application.
For benchmark comparison methods, we used three instance selection methods: random selection (Random), example forgetting (Forgetting)~\cite{toneva2019empirical}, and herding method (Herding)~\cite{chen2010super}.
In addition, we used 10 SOTA dataset distillation methods: dataset condensation~\cite{zhao2021datasetcondensation}, differentiable siamese augmentation (DSA)~\cite{zhao2021differentiatble}, distribution matching (DM)~\cite{zhao2023distribution}, aligning features (CAFE)~\cite{wang2022cafe}, CAFE + DSA~\cite{wang2022cafe}, effective gradient matching (EGM)~\cite{jiang2022delving}, dataset condensation with contrastive signals (DCC)~\cite{lee2022dataset}, matching training trajectories (MTT)~\cite{cazenavette2022dataset}, dataset distillation using parameter pruning (DDPP)~\cite{li2023ddpp}, and kernel inducing point (KIP)~\cite{nguyen2021kipimprovedresults}.
\par
This study presents experimental results, which involve the training of five networks from scratch on the distilled dataset, with the average accuracy and standard deviation reported as performance metrics. 
For the other SOTA methods, we directly used the reported accuracy in the original papers for fair comparison.
We adopted a sample 128-width ConvNet~\cite{gidaris2018dynamic} as the network structure for dataset distillation. 
The architecture of ConvNet comprises multiple convolution blocks, each containing a 3 $\times$ 3 convolution layer with 128 filters, instance normalization, ReLU activation, and 2 $\times$ 2 average pooling with stride two. 
A single linear layer generates logits after the convolution blocks. 
The number of such blocks is determined by the dataset resolution and specified for each benchmark dataset.
To ensure that the distillation process does not collapse when the self-adaptive weights $\mathcal{W}$ approach zero, we set the learning rate of the SGD optimizer to a low value, such as 1e-4, which can control the update speed of the self-adaptive weights $\mathcal{W}$.
\begin{figure}[t]
        \centering
        \includegraphics[width=8cm]{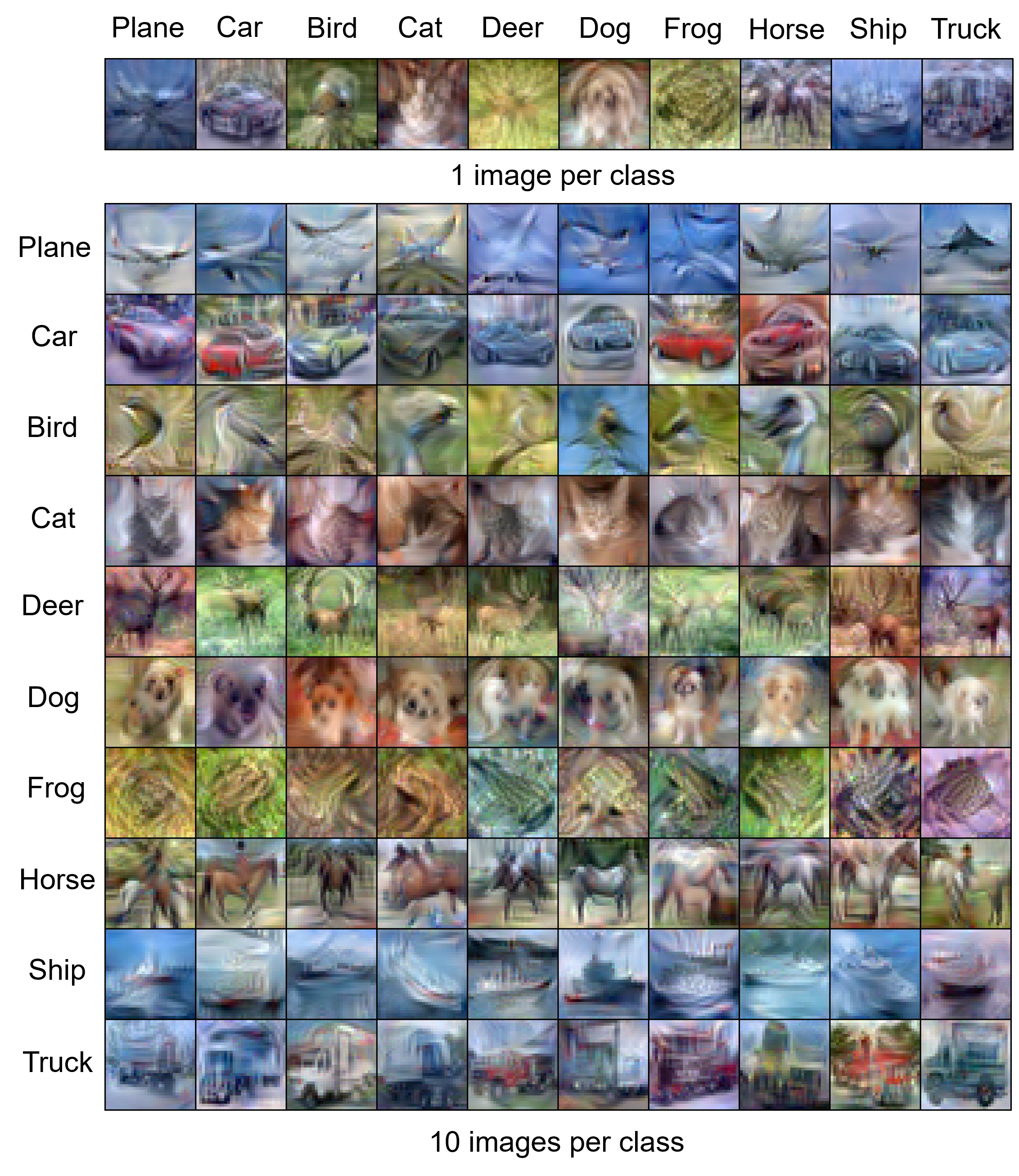}
        \caption{Distilled CIFAR-10 dataset with IPC = 1 and 10.}
        \label{fig3_}
\end{figure}
\begin{figure*}[t]
        \centering
        \subfigure[CIFAR-100]{
        \includegraphics[width=8cm]{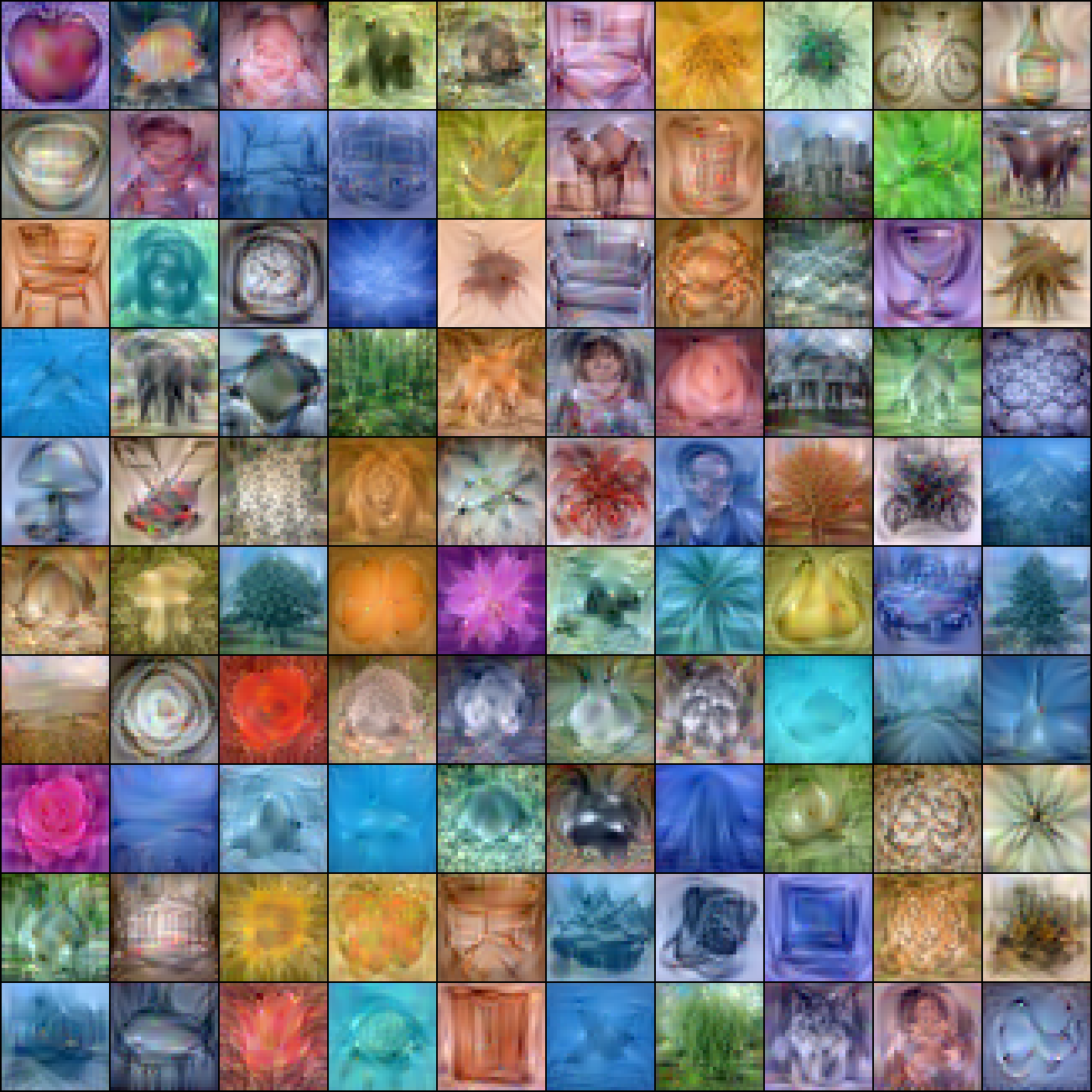}
        }
        \subfigure[Tiny ImageNet]{
        \includegraphics[width=8cm]{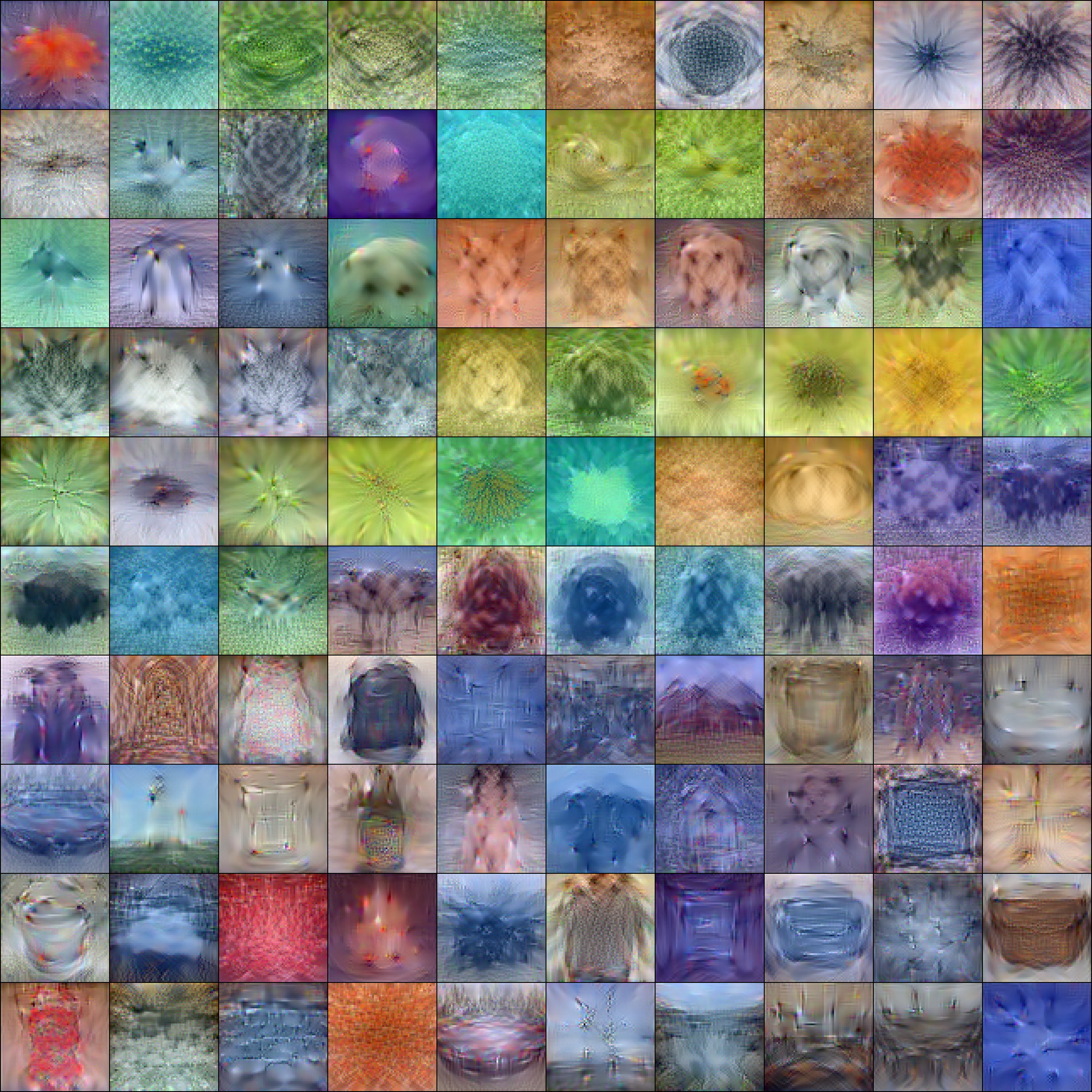}
        }
        \caption{Distilled CIFAR-100 and Tiny ImageNet datasets (selected images) with IPC = 1.}
        \label{fig3}
\end{figure*}
\par
To evaluate the effectiveness of the proposed method in real-world medical applications, we performed experiments using training from scratch (From Scratch) and transfer learning as two baseline methods. 
In addition, we used six SOTA self-supervised learning methods: simple siamese self-supervised learning (SimSiam)~\cite{chen2021exploring}, bootstrap your own latent (BYOL)~\cite{grill2020bootstrap}, cross-view self-supervised learning (Cross)~\cite{li2022covid}, self-knowledge distillation based self-supervised learning (SKD)~\cite{li2022self}, masked autoencoder (MAE)~\cite{he2022masked}, and region-guided masked image modeling (RGMIM)~\cite{li2022rgmim}.
These self-supervised learning methods generally use ResNet50~\cite{he2016deep} or ViT-Base~\cite{dosovitskiy2021image} as their network structures.
The performance of each method was evaluated in terms of four-class accuracy on a large COVID-19 CXR dataset.
\begin{figure*}[ht!]
        \centering
        \subfigure[]{
        \includegraphics[width=8.0cm]{Image/difference_cifar10-1.png}}
        \subfigure[]{
        \includegraphics[width=8.0cm]{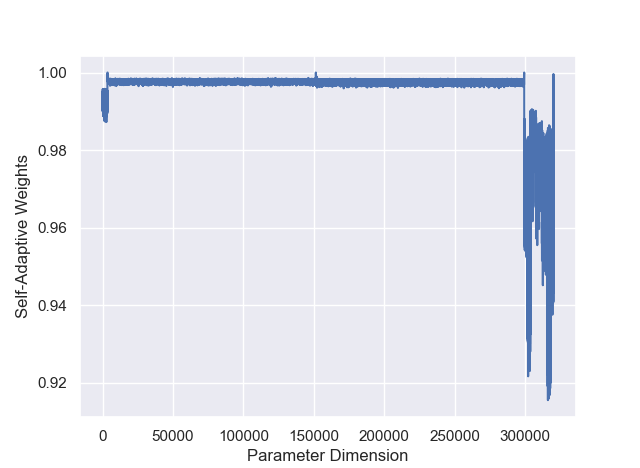}}
        \subfigure[]{
        \includegraphics[width=8.0cm]{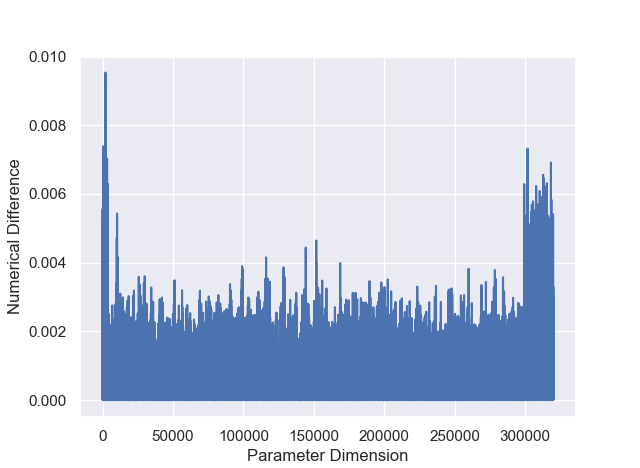}}
        \subfigure[]{
        \includegraphics[width=8.0cm]{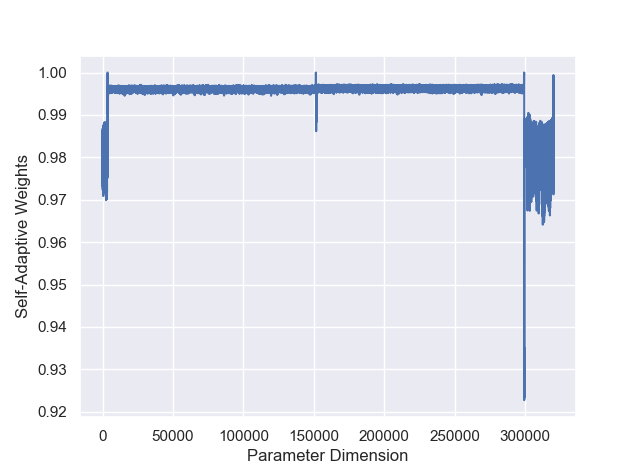}}
        \subfigure[]{
        \includegraphics[width=8.0cm]{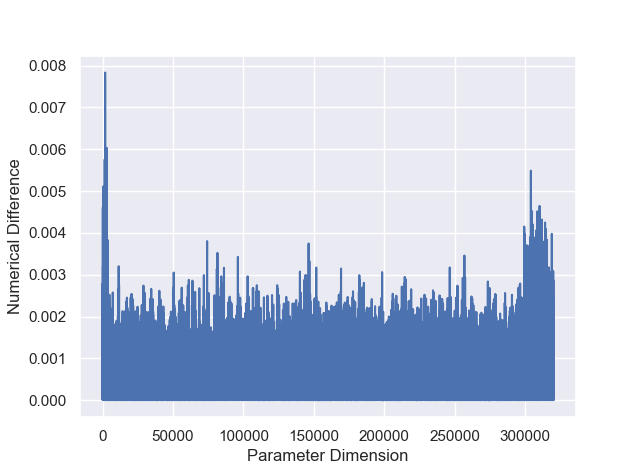}}
        \subfigure[]{
        \includegraphics[width=8.0cm]{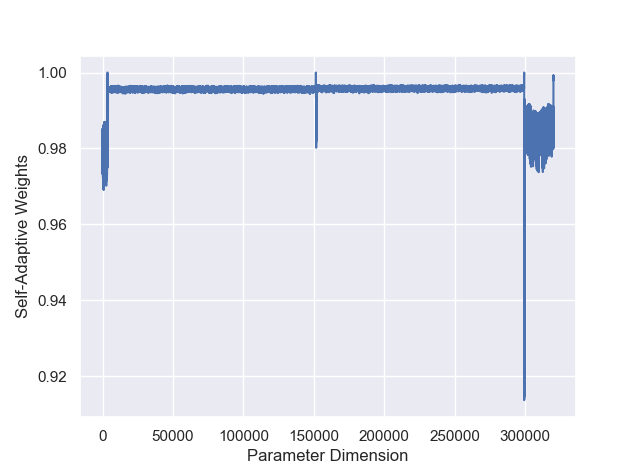}}
        \caption{Analysis of self-adaptive weights of CIFAR-10 with various IPCs. (a), (c), and (e) present the visualization results of the numerical difference between the teacher and student network parameters in the corresponding dimensions with IPC = 1, IPC = 10, and IPC = 50, respectively. (b), (d), and (f) depict the visualization results of the optimized self-adaptive weights in the corresponding dimensions with IPC = 1, IPC = 10, and IPC = 50, respectively.}
        \label{fig4_1}
\end{figure*}
\begin{figure*}[ht!]
        \centering
        \subfigure[]{
        \includegraphics[width=8.0cm]{Image/difference_cifar10-1.png}}
        \subfigure[]{
        \includegraphics[width=8.0cm]{Image/weights_cifar10-1.png}}
        \subfigure[]{
        \includegraphics[width=8.0cm]{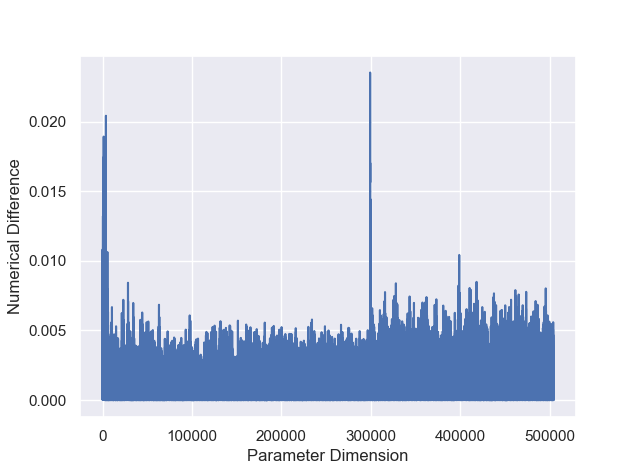}}
        \subfigure[]{
        \includegraphics[width=8.0cm]{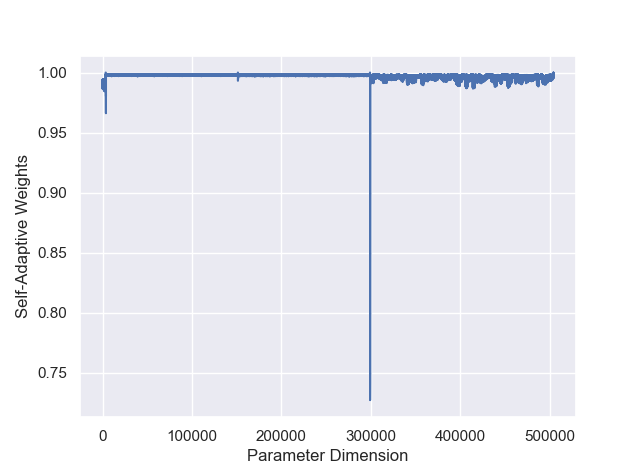}}
        \subfigure[]{
        \includegraphics[width=8.0cm]{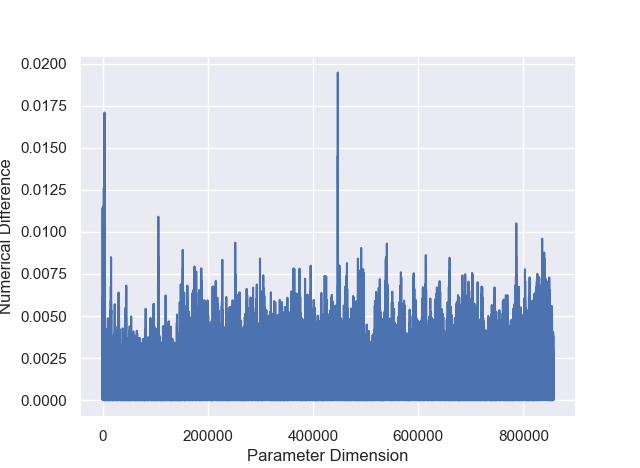}}
        \subfigure[]{
        \includegraphics[width=8.0cm]{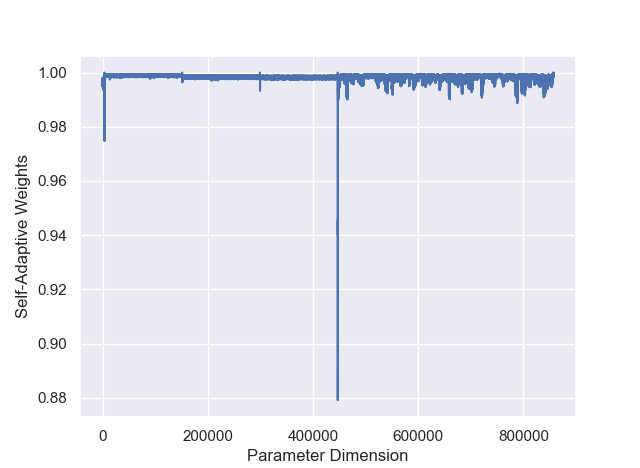}}
        \caption{Analysis of self-adaptive weights of different datasets with IPC = 1. (a), (c), and (e) show the visualization results of the numerical difference between teacher and student network parameters in the corresponding dimensions on CIFAR-10, CIFAR-100, and Tiny ImageNet, respectively. (b), (d), and (f) show the visualization results of the optimized self-adaptive weights in the corresponding dimensions on CIFAR-10, CIFAR-100, and Tiny ImageNet, respectively.}
        \label{fig4_2}
\end{figure*}
\subsection{Benchmark Comparison}
This section presents a comparative analysis of the proposed dataset distillation method against other SOTA methods on three benchmark datasets: CIFAR-10, CIFAR-100, and Tiny ImageNet. 
The proposed method uses zero-phase component analysis (ZCA) whitening with default parameters for distillation performance improvement~\cite{cazenavette2022dataset}.
We used a three-depth ConvNet (ConvNetD3) for CIFAR-10 and CIFAR-100 and a four-depth ConvNet (ConvNetD4) for Tiny ImageNet.
We pretrained 200 teacher networks for CIFAR-10 and CIFAR-100 and 100 teacher networks for Tiny ImageNet, each for 50 epochs. 
IPC denotes the number of distilled images per class.
The number of distillation iterations $T$ was set to 5,000, and the IPC was varied across datasets, with values of 1, 10, and 50 for CIFAR-10 and CIFAR-100 and 1 and 10 for Tiny ImageNet.
The above settings are the same as those in the previous dataset distillation methods.
For a fair comparison with KIP~\cite{nguyen2021kipimprovedresults}, we used their original 1024-width ConvNet (KIP-1024) and 128-width ConvNet (KIP-128).
In addition, we used custom ZCA in the distillation and evaluation processes.
\par
The experimental results in Table~\ref{tab1} reveal the superiority of IADD over the existing dataset selection and dataset distillation methods across most IPC settings.
For CIFAR-10 with IPC = 10, IADD achieved a 1.2\% increase in accuracy compared with our previous DDPP method.
For CIFAR-100 with IPC = 10, IADD achieved a 2.6\% increase in accuracy compared with the SOTA method MTT.
Notably, when all importance weights are uniform, IADD degrades to MTT, indicating the effectiveness of the self-adaptive weights.
DDPP may not be suitable for large-scale datasets because it tends to collapse during the distillation process because of the pruned parameters.
In contrast, the proposed method based on self-adaptive weights is expected to perform efficiently on large-scale datasets, because it can effectively adapt to data characteristics and adjust the importance of different parameters during the distillation process.
\par
Table~\ref{tab2} shows that our method outperforms KIP when using the same 128-width ConvNet.
Even when KIP used a 1024-width ConvNet, IADD still achieved higher accuracy, except for CIFAR-10 with IPC = 1. 
We did not conduct experiments for KIP on CIFAR-100 with IPC = 50 because of the computational resource limitations of KIP; therefore, we only report our results in this study.
The KIP method was specifically designed to function on large-width neural networks in the dataset distillation task. 
However, IADD can achieve suitable distillation performance with a normal-width network structure, which reduces both computing costs and time.
\par
The visualization results of the distilled datasets are shown in Figs.~\ref{fig3_} and~\ref{fig3}, along with CIFAR-10 with IPC = 1 and 10, CIFAR-100 dataset with IPC = 1, and Tiny ImageNet (selected images) with IPC = 1. 
As illustrated in Fig.~\ref{fig3_}, the generated images appear more diverse when IPC = 10, given that the distillation process can compress the discriminative features of a class into multiple images.
For instance, the generated images contain differently colored cars in the second row and various types of horses in the eighth row. 
Furthermore, when IPC = 1, the generated images are more abstract and densely packed with information, as the distillation process requires all class features to be compressed into a single image.
In addition, when the resolution increases (from CIFAR-100 to Tiny ImageNet), as depicted in Fig.~\ref{fig3}, the texture of the generated image is clearer and can show certain details, such as the beak of a penguin in the second image from the left in the third row and the ears of a cat in the third image from the left in the fourth row.
The distilled images of the other SOTA methods can be easily found in their original papers.
\par
We also conducted runtime complexity analysis of MTT, DDPP, and IADD. 
We measured the runtime results over 10 iterations, with each iteration consisting of 50 matching steps. 
These computations were performed on a single NVIDIA RTX A6000 GPU, with a batch size of 100.
The results reveal that the distillation runtimes of MTT and DDPP average 11.5±0.5 and 12.5±0.5 s, respectively.
In contrast, IADD exhibits a slightly higher distillation runtime, averaging 13.0±0.5 s. 
These findings indicate that the optimization of self-adaptive weights in our approach does not considerably increase the distillation runtime.
\subsection{Analysis of Self-Adaptive Weights}
In this section, we present two experiments to analyze the self-adaptive weights of IADD from different perspectives.
The self-adaptive weights $\mathcal{W}$ are initialized with $\mathcal{W}_{0}$, a unit vector with corresponding dimensions. 
Subsequently, we proceeded with the distillation process and iteratively optimized the self-adaptive weights $\mathcal{W}$ via SGD until convergence was achieved.
\par
In the first experiment, we demonstrated the analysis of the self-adaptive weights of CIFAR-10 with different IPC values.
Figures~\ref{fig4_1} (a), (c), and (e) show the visualization results of the numerical difference between the parameters of the teacher and student networks in the corresponding dimensions with IPC = 1, IPC = 10, and IPC = 50, respectively.
Figures~\ref{fig4_1} (b), (d), and (f) show the visualization results of the optimized self-adaptive weights in the corresponding dimensions with IPC = 1, IPC = 10, and IPC = 50, respectively.
As illustrated in Fig.~\ref{fig4_1}, across all settings, the adaptive weights of parameters with considerable differences gradually decrease during the training process. 
Conversely, the weights of parameters with moderate differences remain high.
The experimental findings indicate that the proposed method for optimizing self-adaptive weights successfully issues the challenges of capturing variations in the complexity of parameter matching across various dimensions. 
This is because the self-adaptive weights can capture the subtle differences in the parameters of the teacher and student networks, and accordingly adjust the relative importance of various parameters to better align the datasets.
In particular, they enable the dataset distillation process to make precise and context-dependent adjustments, thereby improving the distillation performance.
\par
The teacher and student network parameters may exhibit non-uniformity across the corresponding dimensions, posing a challenge in matching these parameters. 
This non-uniformity is particularly prominent in the early and late layers, corresponding to input and output processing, respectively.
These layers serve distinct functions, such as feature extraction and classification, and are thus subject to different sources of variation. 
Consequently, achieving an optimal balance of the parameters in these layers is critical for realizing high distillation performance. 
The proposed method for optimizing self-adaptive weights successfully overcomes these challenges by enabling precise and context-dependent adjustment of network parameters, resulting in improved distillation performance.
\par
In the second experiment, we present the analysis of the self-adaptive weights of different datasets with IPC = 1.
Figures~\ref{fig4_2} (a), (c), and (e) show the visualization results of the numerical difference between teacher and student network parameters in the corresponding dimensions on CIFAR-10, CIFAR-100, and Tiny ImageNet, respectively. 
Figures~\ref{fig4_2} (b), (d), and (f) show the visualization results of the optimized self-adaptive weights in the corresponding dimensions on CIFAR-10, CIFAR-100, and Tiny ImageNet, respectively.
As illustrated in Fig.~\ref{fig4_2}, across all settings, we observed that the parameters exhibiting considerable disparities tend to experience a gradual reduction in their weight values, whereas those with moderate differences maintain higher weights, which is consistent with the previous experiment.
\par
In addition, we found that the optimization process of the self-adaptive weights experienced variations when using different datasets and training with networks of different depths.
Different datasets have unique characteristics in terms of data distribution, feature representation, and noise levels.
Consequently, these variations necessitate adaptive adjustments during the optimization process to effectively capture dataset-specific patterns.
In addition, the depths of neural networks affect the optimization process. 
Networks with varying depths exhibit distinct levels of abstraction and feature hierarchies. 
Shallow networks tend to focus on capturing simpler features, whereas deep networks delve into intricate and abstract representations. 
These disparities influence the matching and weighing parameters in the optimization process. 
Deeper layers typically involve more abstract and task-specific information affecting the matching dynamics and consequently influencing the weight assignment.
\subsection{Cross-Architecture Generalization}
This section presents an evaluation of the cross-architecture generalization capability of the proposed method, where distilled images generated by ConvNetD3 on CIFAR-10 are used for testing on other architectures.
The IPC was set to 10, and the same pretrained teacher networks used in Section 4.2 were used for rapid distillation and experimentation.
The three main networks, namely AlexNet~\cite{krizhevsky2012imagenet}, VGG11~\cite{simonyan2015very}, and ResNet18~\cite{he2016deep}, were used to evaluate cross-architecture generalization with high accuracy.
\par
Table~\ref{tab3} shows that the proposed method outperforms the SOTA methods MTT and DDPP for all architectures, indicating that our method can generate more robust distilled images.
Notably, our method achieved an accuracy increase of 3.1\% compared with DDPP for ResNet18.
This improvement can be attributed to the automatic assignment of importance weights to various parameters during distillation, improving the contribution of important parameters and penalizing unimportant parameters.
Consequently, our method effectively captures the important features of the data, resulting in improved performance of cross-architecture generalization.
\begin{table}[t]
    \centering
    \caption{Test accuracy of cross-architecture generalization  on CIFAR-10 dataset with IPC = 10.}
    \label{tab3}
    \begin{tabular}{lcccc}
    \hline
    Method & ConvNetD3 & AlexNet & VGG11 & ResNet18 \\\hline\hline
    DSA~\cite{zhao2021differentiatble} & 52.1$\pm$0.4 & 35.9$\pm$1.3 & 43.2$\pm$0.5 & 42.8$\pm$1.0 \\
    KIP~\cite{nguyen2021kipimprovedresults} & 47.6$\pm$0.9 & 24.4$\pm$3.9 & 42.1$\pm$0.4 & 36.8$\pm$1.0 \\
    MTT~\cite{cazenavette2022dataset} & 64.3$\pm$0.7 & 34.2$\pm$2.6 & 50.3$\pm$0.8 & 46.4$\pm$0.6 \\
    DDPP~\cite{li2023ddpp} & 65.4$\pm$0.4 & 35.8$\pm$1.3 & 52.9$\pm$0.9 & 51.8$\pm$1.1 \\
    IADD & \bfseries{66.2$\pm$0.6} & \bfseries{36.5$\pm$1.1} & \bfseries{53.4$\pm$1.3} & \bfseries{54.9$\pm$0.7} \\
    \hline
    \end{tabular}
\end{table}
\begin{table}[t]
    \centering
    \caption{COVID-19 detection accuracy of different methods. The first four comparison methods are baseline methods, the next six comparison methods are SOTA self-supervised learning methods, and the others are SOTA dataset distillation methods.}
    \label{tab4}
    \begin{tabular}{lccc}
    \hline
    Method & Images & Structure & Accuracy \\\hline\hline
    From Scratch & 168 & ResNet50~\cite{he2016deep} & 28.4\% \\
    From Scratch & 168 & ViT-Base~\cite{dosovitskiy2021image} & 41.3\% \\
    Transfer & 168 & ResNet50 & 53.9\% \\
    Transfer & 168 & ViT-Base & 68.9\% \\\hline\hline
    SimSiam~\cite{chen2021exploring} & 168 & ResNet50 & 62.3\% \\
    BYOL~\cite{grill2020bootstrap} & 168 & ResNet50 & 68.3\% \\
    Cross~\cite{li2022covid} & 168 & ResNet50 & 74.7\% \\
    SKD~\cite{li2022self} & 168 & ResNet50 & 74.2\% \\
    MAE~\cite{he2022masked} & 168 & ViT-Base & 75.4\% \\
    RGMIM~\cite{li2022rgmim} & 168 & ViT-Base & 77.1\% \\\hline\hline
    MTT~\cite{cazenavette2022dataset} & 80 & ConvNetD5 & 82.7\% \\
    DDPP~\cite{li2023ddpp} & 80 & ConvNetD5 & 84.1\% \\
    IADD & 80 & ConvNetD5 & \bfseries{85.2\%} \\\hline\hline
    Original Dataset & 16,933 & ConvNetD5 & 88.9\% \\
    \hline
    \end{tabular}
\end{table}
\begin{figure*}[t]
        \centering
        \includegraphics[width=12.5cm]{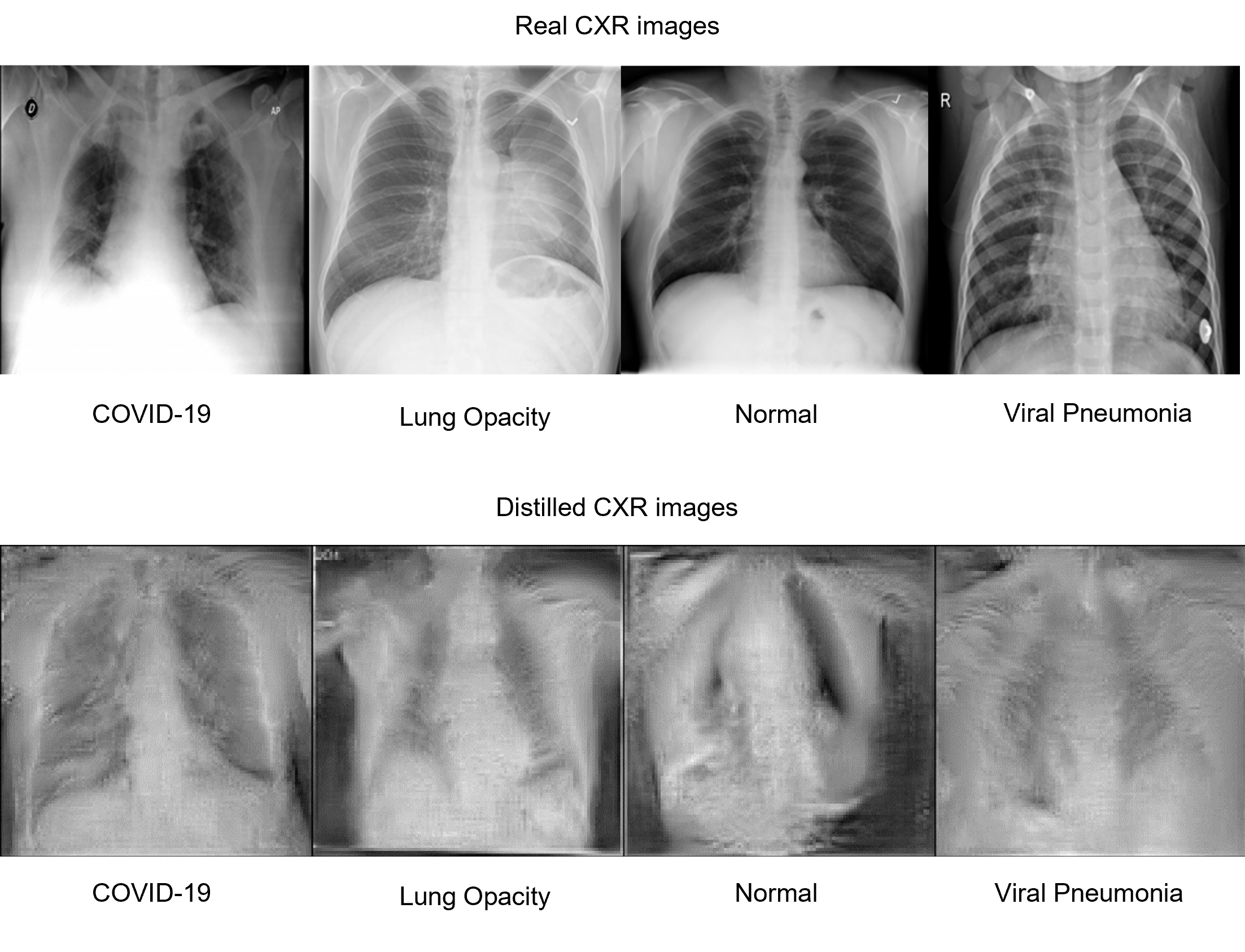}
        \caption{Examples of real and distilled CXR images of four classes (COVID-19, Lung Opacity, Normal, and Viral Pneumonia).}
        \label{fig5}
\end{figure*}
\subsection{Real-World Medical Application}
In this section, the effectiveness of the proposed method is evaluated on a COVID-19 CXR dataset for real-world applications. 
A five-depth ConvNet (ConvNetD5) was used for distillation, which was necessary because of the considerable increase in image resolution compared with CIFAR-10 and CIFAR-100.
Herein, 100 teacher networks were pretrained for the distillation process, with each teacher network trained for 50 epochs. 
The COVID-19 detection accuracy was tested with the IPC set to 20. 
For comparison, the COVID-19 detection accuracy of SOTA self-supervised learning methods was evaluated when 42 randomly selected images per class were chosen, accounting for 1\% of the training set.
For MTT and DDPP, we used the same training settings as those used in the proposed method.
\par
Table~\ref{tab4} shows the superior performance of the proposed method over MTT and DDPP in terms of COVID-19 detection accuracy, particularly when using a small number of distilled CXR images. 
The proposed method achieves high test accuracy even when the IPC value is low (20; corresponding to 80 distilled CXR images). 
Moreover, the proposed method considerably outperforms the SOTA self-supervised learning method, using a simpler network and fewer training images, indicating its effectiveness in real-world COVID-19 detection applications.
In addition, the table presents the upper bound accuracy of 88.9\% when training on the original dataset.
Even with high-level compression, no considerable accuracy degradation is observed, which further demonstrates the effectiveness of the proposed method.
\par
Examples of real and distilled CXR images are shown in Fig.~\ref{fig5}. 
The distilled images exhibit visual differences compared with the original CXR images.
The sharing of medical datasets between hospitals is typically hindered by privacy-protection issues and the high cost of transmitting and storing numerous high-resolution medical images~\cite{ye2021management}. 
Dataset distillation can address this issue by synthesizing a small anonymous dataset such that models trained on it can achieve performance comparable to that when using the original large dataset. 
This potential demonstrates the capability of dataset distillation to solve existing medical sharing problems~\cite{dankar2017risk}.
In addition, the combination of medical dataset distillation and federated learning~\cite{song2023resfed, song2023fedbevt} may contribute to clinical usage.
\section{Discussion}
We have found that certain parameters can be difficult to match during the dataset distillation process, resulting in a negative impact on distillation performance, as observed in our previous work DDPP~\cite{li2023ddpp}. 
To solve this problem, DDPP uses parameter pruning to remove these parameters during the distillation process.
However, directly removing difficult-to-match parameters is ineffective and can affect the number of network model parameters. 
Hence, DDPP is unsuitable for large-scale datasets because it tends to collapse during the distillation process, which is caused by the pruned parameters. 
In contrast to directly removing difficult-to-match parameters, IADD can improve distillation performance by automatically assigning importance weights to parameters during distillation.
Furthermore, because IADD automatically optimizes the weights according to the importance of the parameters, we do not encounter the hyperparameter tuning problem in DDPP for defining difficult-to-match parameters.
The promising results provide further evidence that supports the potential of the proposed method to contribute to the development of more efficient and effective dataset distillation algorithms.
\par
For the other SOTA dataset distillation methods, DM~\cite{zhao2023distribution} focuses on aligning dataset distributions to improve parameter alignment. 
Its efficacy depends on the specific distribution matching strategy employed.
CAFE~\cite{wang2022cafe} emphasizes feature-level alignment through feature matching. 
Its performance is influenced by factors such as feature complexity and network architecture.
MTT~\cite{cazenavette2022dataset} concentrates on aligning the training process for dynamic knowledge transfer. 
This method necessitates meticulous tuning and careful consideration of difficult-to-match parameters.
KIP~\cite{nguyen2021kipimprovedresults} was specifically designed to operate on large-width neural networks in the dataset distillation task.
In contrast, our method, IADD, achieves suitable distillation performance with a normal-width network structure. 
This distinction reduces the computing costs and considerably lowers the processing time.
\par
Although our proposed method has shown promising results in parameter matching-based dataset distillation, it has several limitations. 
First, the proposed method is specifically designed for parameter matching-based algorithms, and extending it to other types of dataset distillation such as meta-learning or distribution matching remains to be researched further. 
Second, the proposed method may not apply to large-scale models such as vision transformers because of their high number of parameters, which can impact the optimization of self-adaptive weights.
Moreover, although the proposed method has demonstrated its effectiveness in COVID-19 detection, its performance on other downstream tasks, such as continuous learning or neural network architecture search, is yet to be validated. 
\par
We also highlight some possible future research directions.
Existing dataset distillation methods mainly rely on supervised learning, and the potential of unsupervised or self-supervised learning-based methods for dataset distillation needs to be investigated~\cite{liu2021self}.
According to the results from the COVID-19 dataset, we do not observe an explicit effect of the imbalance on the distillation process, probably because the imbalance ratio is not very high. 
The dataset imbalance or long-tail problem is an interesting topic for dataset distillation, and it will be one of our future research directions~\cite{zhang2023deep}.
\section{Conclusion}
In this study, we propose a novel dataset distillation method called IADD that improves distillation performance by assigning importance weights to different network parameters during distillation and generating more robust distilled datasets. 
Specifically, the proposed method improves the contribution of important parameters and penalizes unimportant parameters during the distillation process. 
Our experimental results show that IADD outperforms other SOTA distillation methods based on parameter matching on three benchmark datasets and exhibits improved cross-architecture generalization performance.
Furthermore, we validate the effectiveness of our method on a real-world COVID-19 CXR dataset.
\par
Our method has shown promise in parameter matching-based dataset distillation; however, it has some limitations. 
It is specifically tailored for parameter matching-based methods and may not easily extend to other dataset distillation methods such as meta-learning or distribution matching. 
Furthermore, its applicability to large-scale models such as vision transformers remains uncertain because of potential challenges in optimizing self-adaptive weights for such models.
\par
Given that existing dataset distillation methods predominantly rely on supervised learning paradigms, there is a compelling need to delve into the uncharted territory of unsupervised or self-supervised learning-based methods for dataset distillation.
As part of our ongoing research, we intend to delve deeper into the intricacies of dataset imbalance and its implications for dataset distillation, particularly in scenarios characterized by significant class imbalances.
\section*{Ethical Approval}
No ethics approval is required.
\section*{Declaration of Competing Interest}
None declared.
\section*{Acknowledgments}
This study was partly supported by JSPS KAKENHI Grant Numbers JP21H03456 and JP23K11141, and AMED Grant Number JP23zf01270004h0003. This study was conducted at the Data Science Computing System of Education and Research Center for Mathematical and Data Science, Hokkaido University.
\bibliographystyle{elsarticle-num}
\bibliography{NN}
\end{document}